\documentclass[journal]{IEEEtran}

\ifCLASSINFOpdf
\else
   \usepackage[dvips]{graphicx}
\fi
\usepackage{url}

\usepackage{booktabs} 
\usepackage{graphicx}
\usepackage{caption}
\usepackage{multirow}
\usepackage{subcaption}
\usepackage{hyperref}
\usepackage{amsmath}

\usepackage{paralist}

\usepackage{flushend}

\usepackage[numbers,sort&compress]{natbib}

\usepackage[numbers]{natbib} 
\hypersetup{
	colorlinks=true,
	linkcolor=black,
	citecolor=black,
	urlcolor=black,
	citecolor=black,
}
\begin{document}

\title{\fontsize{23pt}{30pt}\selectfont A Pairwise DomMix Attentive Adversarial Network for Unsupervised Domain Adaptive Object Detection}

\author{{Jie Shao}, {Jiacheng Wu}, {Wenzhong Shen}, {Cheng Yang}.
\thanks{The authors are with the Department of Electronics and information engineering, Shanghai University of Electric Power, Shanghai 201306, China (e-mail: shaojie@shiep.edu.cn; 657685808@mail.shiep.edu.cn; shenwenzhong@shiep.edu.cn; cheng.yang@ shiep.edu.cn).}}

\markboth{Journal of \LaTeX\ Class Files, Vol. 14, No. 8, August 2015}
{Shell \MakeLowercase{\textit{et al.}}: Bare Demo of IEEEtran.cls for IEEE Journals}
\maketitle

\begin{abstract}
Unsupervised Domain Adaptive Object Detection (DAOD) could adapt a model trained on a source domain to an unlabeled target domain for object detection. Existing unsupervised DAOD methods usually perform feature alignments from the target to the source. Unidirectional domain transfer would omit information about the target samples and result in suboptimal adaptation when there are large domain shifts. Therefore, we propose a pairwise attentive adversarial network with a Domain Mixup (DomMix) module to mitigate the aforementioned challenges. Specifically, a deep-level mixup is employed to construct an intermediate domain that allows features from both domains to share their differences. Then a pairwise attentive adversarial network is applied with attentive encoding on both image-level and instance-level features at different scales and optimizes domain alignment by adversarial learning. This allows the network to focus on regions with disparate contextual information and learn their similarities between different domains. Extensive experiments are conducted on several benchmark datasets, demonstrating the superiority of our proposed method.

\end{abstract}


\IEEEpeerreviewmaketitle

\section{Introduction}

\IEEEPARstart{D}{omain} adaptation for object detection (DAOD) is a challenging task that involves adapting object detectors from a source domain to a target domain with significant differences. A lot of early methods required labeled data in both domains\cite{saito2019strong,chen2020harmonizing,deng2021unbiased,motiian2017unified}, whose practicability is limited by repeated label annotation and diversity of target samples. Therefore, unsupervised DAOD has gained considerable attention recently.

Methods for unsupervised DAOD can be roughly divided into two categories. The first category is based on context-related transfer learning, such as graph matching\cite{li2022sigma,vs2022instance,li2021synthetic}, instance-level feature aggregation\cite{shen2019scl,deng2022text,an2022effectiveness,qian2022successive}, etc.. They rely on distance minimization of feature distributions between different domains, which are usually applied to images with prominent style differences, but suffer from pseudo-labeling noises and inflexible distance metrics. Another category employs adversarial training\cite{ganin2015unsupervised,tzeng2017adversarial} to discriminate domain mismatch and alleviate domain shift. In contrast to the first category, it could flexibly implement feature alignments between two different domains during training, but perform poorly when there are large domain shifts. Recent studies have attempted to help the discriminator find correlations between domains by generating mixed samples using methods related to data augmentation, such as CutMix\cite{yun2019cutmix}, Mixup\cite{zhang2017mixup}, Augmix\cite{hendrycks2019augmix} and Confmix\cite{mattolin2023confmix} etc.. However, pixel-level fusion based on data augmentation will reduce recognition accuracy in the source domain and lead to loss of information in the target domain as well. Later, SPANet\cite{li2020spatial} focused on domain differences by applying attention, but it treated the image-level and instance-level features the same, which resulted in information redundancy and overfitting to the source. Above all, although great efforts have been made on unsupervised DAOD, feature alignment with large domain shifts is still a challenging issue.

Targeting this problem, we propose a pairwise DomMix attentive adversarial network (PDAANet) for unsupervised DAOD. First of all, inspired by the mixup strategy in data augmentation, we construct an intermediate domain that lies between source and target with deep multi-scale image-level feature mixup, and hence avoid unidirectional mapping across two distributions with a significant gap. Specifically, multi-scale image-level features from the two domains are pairwise mixed and transformed to form a group of intermediate features with hybrid distributions, which is an easier task than directly aligning the target to the source. We call it the DomMix module. After initiating the intermediate domain with DomMix, instance-level features and multi-scale mixed image-level features are put into a pairwise attentive module. It pays attention to the image-level and the instance-level features respectively, and especially embeds a Residual SimAM\cite{yang2021simam} structure to focus on the localization information in cross-domain object detection. Later, the generated aligned features are discriminated between the source and the target with an adaptive pyramid classifier. Finally, loss combinations are proposed to facilitate cross-domain transfer. Our PDAANet shows excellent performance in cross-domain detection and mitigates the shortcomings of existing prototype-based domain adaptation methods. Overall, our contributions are as follows:
\begin{itemize}
	\item[1)] We propose a pairwise DomMix attentive adversarial learning framework, which constructs an intermediate domain by bidirectional mapping, serving as an alignment bridge for source and target domains with large domain shifts.
	\item[2)]  Different from most attention strategies that are more suitable for single-domain features, a  Residual SimAM (RSA) structure is specially designed for attentive encoding on cross-domain target localization.
	\item[3)] Our method achieves state-of-the-art performance on domain adaptation benchmarks through extensive experiments, demonstrating the effectiveness of the proposed method.
\end{itemize}
\begin{figure*}
	\centerline{\includegraphics[width=0.9\linewidth]{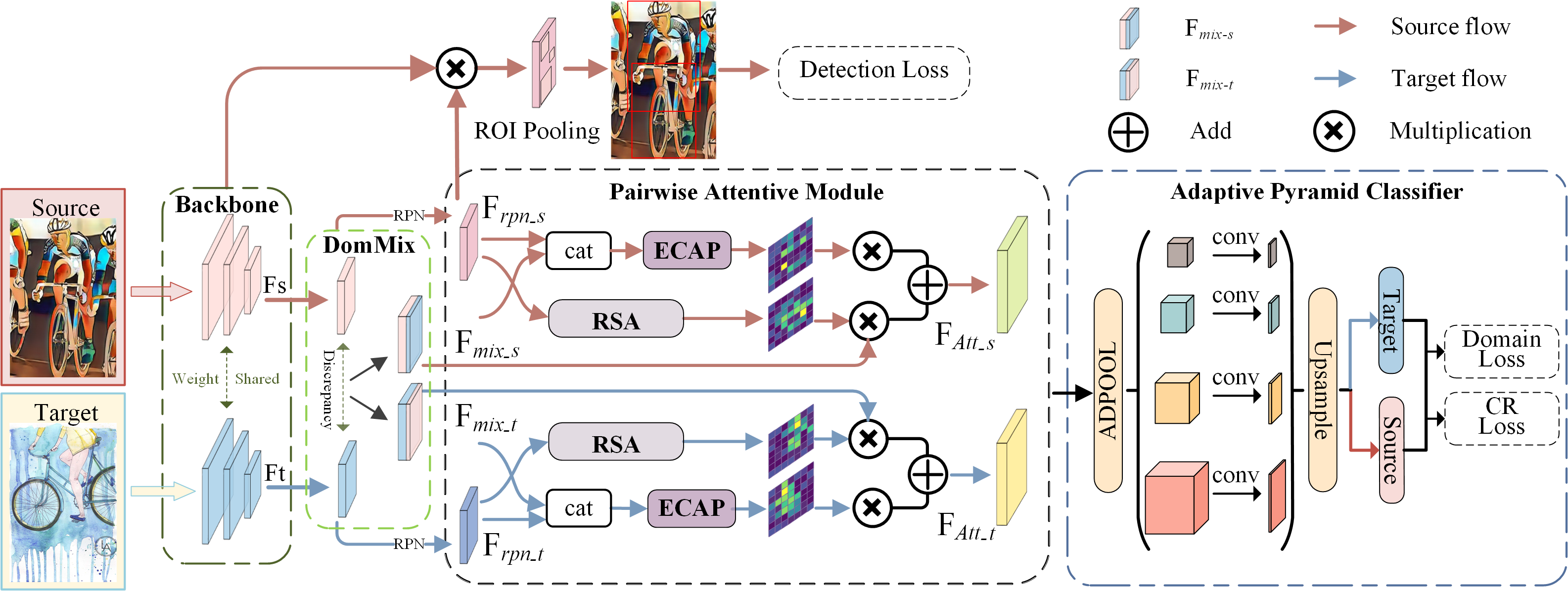}}
	\caption{The overview of our method.}\label{fig2}
\end{figure*}

\section{APPROACH}
\subsection{Overview}
Fig. 1 shows the overview of our proposed PDAANet method. It consists of three main stages: intermediate domain construction by DomMix and PAM; discrimination by an adaptive pyramid classifier and loss optimization; cross-domain object detection after adversarial learning. First, source and target domain $D_s=\{(x_s^{i},y_s^i)\}_{i=1}^M$ and $D_t=\{(x_t^j)\}_{j=1}^N$ are from inconsistent domain distributions $P_s$ and $P_t$. They are followed by the backbone for feature extraction, resulting in $F_s=\{f_{s(k)}^i\}$ and $F_t=\{f_{t(k)}^j\}$, $k \in \{2,3\}$ respectively, where k is the layer index. Then $F_s$ and $F_t$ are put into RPN for instance-level features $F_{rpn}=\{F_{rpn\rule{.3em}{.3pt}s}, F_{rpn\rule{.3em}{.3pt}t}\}$. Meanwhile, they are mixed mapped in an intermediate domain to generate $F_{mix}=\{F_{mix\rule{.3em}{.3pt}s}, F_{mix\rule{.3em}{.3pt}t}\}$ by DomMix. Subsequently, $\{{F_{rpn\rule{.3em}{.3pt}\theta}, F_{mix\rule{.3em}{.3pt}\theta}\}}$, $\theta\in \{s,t\}$ is put into the PAM with two separate branches to obtain attention-weighted feature maps $F_{Att}^s$ and $F_{Att}^t$ for the source and target samples, respectively, which are then used as inputs for the adaptive pyramid classifier. Later, the APC is used as the domain discriminator for adversarial training. Finally, cross-domain detection is performed by the aligned features after training.

\vspace{-5pt} 
\subsection{The DomMix Module}
We propose a domain mixup (DomMix) strategy to establish an intermediate domain, serving as a bridge between source and target domains. It includes the source dominant mappings and the target dominant mappings. Specifically, $F_t=\{f_{t(k)}^j\}$ are initially extracted from samples in the target domain and then integrated into $F_s=\{f_{s(k)}^i\}$ as noise to accomplish source-dominate mapping $F_{mix\rule{.3em}{.3pt}s(k)}^{ij}$. It serves as a form of perturbation that brings variation to the source, so as to improve the variability and adaptability of the model. At the same time, $F_s=\{f_{s(k)}^i\}$ extracted from samples in the source domain are incorporated into $F_t=\{f_{t(k)}^j\}$ to align features between source and target domain samples during training in the target-dominant mapping. Given a pair of input samples $x_s^i$ and $x_t^{j}$ from source and target. Our DomMix settings are defined as follows:

\begin{equation}\label{eqn-1} 
	\begin{aligned}
		F_{mix\rule{.3em}{.3pt}s(k)}^{ij} = \lambda F_{s(k)}^i+(1-\lambda) F_{t(k)}^j \\
		F_{mix\rule{.3em}{.3pt}t(k)}^{ji} = \lambda F_{t(k)}^j+(1-\lambda) F_{s(k)}^i
	\end{aligned}
\end{equation}
where $\lambda \in [0,1]$. The DomMix module provides a clear and controllable mixing mechanism, whose outputs are employed as the initial value of the intermediate domain.

\vspace{-10pt} 
\subsection{The Pairwise Attentive Module}
The pairwise attentive module is proposed following the DomMix to capture the localization-relevant features and reduce the domain discrepancy by suppressing the disparate domain-specific features. It guides the model to find context cues from samples in different domains. Different from existing attention mechanisms \cite{hu2018squeeze, woo2018cbam} that refine features along either channel or spatial dimensions and are not suitable for region proposals in DAOD, our PAM could focus on positioning features at different scales and weigh region proposals. Specifically, it quantifies the importance of RPN regions with mixed multi-scale image-level features and investigates correlations between RPN features and the task of cross-domain object detection. It consists of two branches. One branch is applied to the source-dominant mappings and the source RPN features, and the other is applied to the target-dominant mappings and the target RPN features. Each branch is composed of two parallel branchlets. One branchlet is Residual SimAM (RSA), which is specifically designed for region proposals, as shown in Fig. 2. The other is called Effective Channel Attention Plus (ECAP).

The RSA is inspired by SimAM, which has been proven to be flexible and effective in feature representation. However, SimAM performs poorly in unsupervised DAOD because domain shift would destroy its energy distribution and the energy operator and could not handle the diversity of domain distributions. Different from SimAM, we first apply a convolutional layer in RSA to extract deeper features and then calculate their cosine similarities to generate 3D weights of the features. Therefore, only the directional similarity between RPN features is measured, while domain shifts about the magnitude are actively ignored. Besides, a residual structure is added to improve the perception of high-frequency information in the spatial domain. Specifically, RSA computes the global attention weights of $F_{rpn}$ and applies them to its corresponding elements of $F_{mix}$ to obtain a hybrid feature after mapping the RPN weights by the dot product. It improves attention on targets in the RPN region and thus enhances the perception of targets by hybrid features. The RSA could be calculated as follows:
\vspace{-5pt} 
\begin{equation}\label{eqn-2} 
	att_{RSA\rule{.3em}{.3pt}\theta}=RSA(F_{rpn\rule{.3em}{.3pt}\theta})\cdotp F_{mix\rule{.3em}{.3pt}\theta},\theta\in \{s,t\}
\end{equation}
\vspace{-20pt} 
\begin{figure}[h]
	\begin{minipage}[t]{0.42\linewidth}
		\centering
		\includegraphics[scale=1.43]{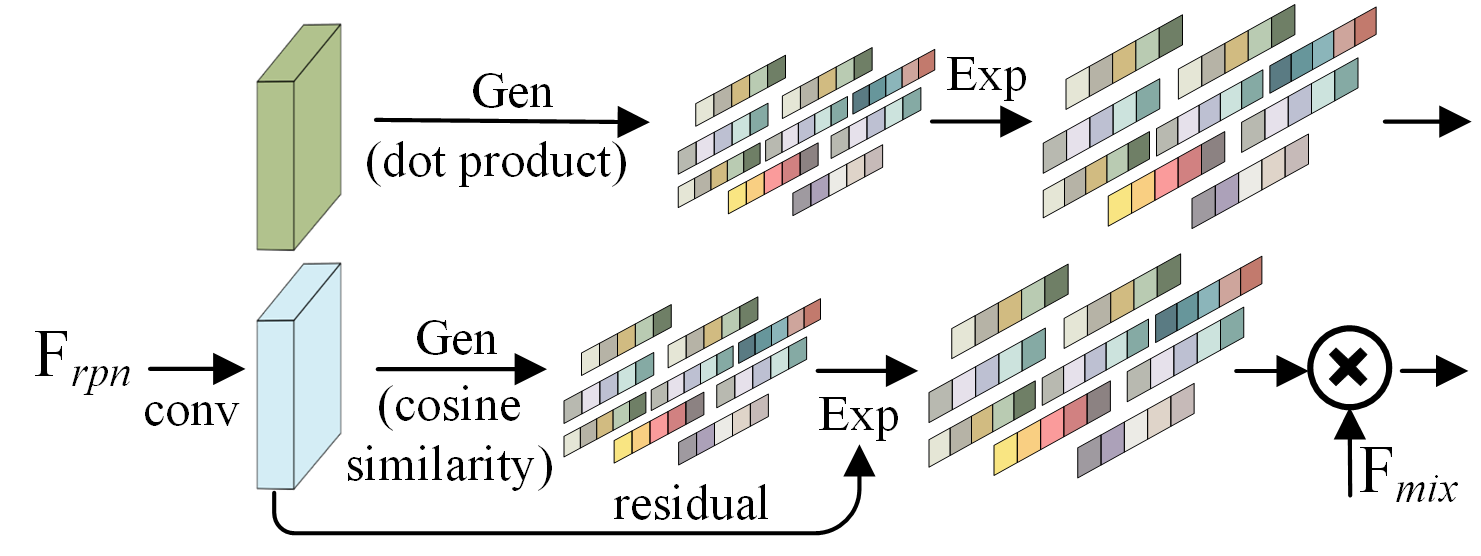}
	\end{minipage}%
	\captionsetup{font={small}}
	\caption{SimAM (up) and RSA (bottom)}
\end{figure}
\vspace{-5pt} 

On the other hand, the ECAP branchlet is performed after the combination of $F_{rpn}$ and $F_{mix}$. We extend the Effective Channel Attention (ECA)\cite{wang2020eca} by adding a 1×1 convolution layer, which serves as a channel-wise weighting factor to adaptively adjust the weights of ECA and alleviate the risk of over-fitting on the source. The final output $F_{Att}$ is the weighted combination of RSA and ECAP. The ECAP branchlet and the $F_{Att}$ can be represented as:
\vspace{-5pt} 
\begin{align}\label{eqn-3} 
	att_{ECAP\rule{.3em}{.3pt}\theta}=ECAP(F_{rpn\rule{.3em}{.3pt}\theta},F_{mix\rule{.3em}{.3pt}\theta}), \theta\in \{s,t\}  \\
	F_{Att\rule{.3em}{.3pt}\theta}=\alpha\cdotp att_{RSA\rule{.3em}{.3pt}\theta}+(1-\alpha)\cdotp att_{ECAP\rule{.3em}{.3pt}\theta}, \theta\in \{s,t\}
\end{align}
where $\alpha$ is the attentional weighting factor.

\subsection{Adversarial Learning}
Following the pairwise attentive module, an Adaptive Pyramid Classifier (APC) is proposed as a domain discriminator in adversarial learning. The discriminator aims to discriminate where the mixed features of the intermediate domain come from (source or target).

The APC is inspired by the Pyramid Pooling Module, which tends to underperform in cross-domain detection tasks due to limitations posed by fixed-scale pooling operations. Therefore, our APC incorporates adaptive pooling layers that automatically adjust pooling operations based on the size and content of the input feature maps, allowing flexible adaptation to contextual information of different scales and shapes. It decomposes the traditional pooling operation into four pooling layers of different scales: 1x1, 2x2, 3x3, and 6x6. These four pooling layers downsample the input feature maps to the target size, which is then restored to the original input size using upsampling operation. The restored feature maps are cascaded into a unified size and used for linear layer classification operations.

\subsection{Full Objective and Inference}
The overall loss $L$ comprises detection loss $L_{det}$, domain loss $L_{dom}$, and consistency regularization loss $L_{cr}$, which is summarized as follows:
\vspace{-5pt} 
\begin{equation}\label{eqn-5} 
	L=\mathop {max} \limits_{D}\mathop {min} \limits_{G,R} (L_{det}+\lambda_1L_{cr}-\lambda_2L_{dom})
\end{equation}
where $\lambda_1, \lambda_2$ controls the trade-off between losses, the detection loss $L_{det}$ is based on the Faster R-CNN for object detection in the source domain, and the consistency regularization loss $L_{cr}$ ensures stable convergence of the two domains during training. Further details of $L_{det}$ and $L_{cr}$ please refer to the original paper\cite{ren2015faster, laine2016temporal}.

The domain loss $L_{dom}$ is used to establish a decision boundary for discrimination between source and target domains. This is achieved by penalizing the discrepancies between the domain-specific features extracted from the intermediate layers of the network. Specifically, we calculate the probability of the sample $x_\theta^i$ belonging to the target domain. The proposed PDAANet is denoted as $P$. The domain classification loss is then calculated as:
\begin{equation}\label{eqn-6} 
	L_{dom}=\frac{1}{\left| D_s\cup D_t \right|}\sum_{x\in D_s\cup D_t}\mathcal F(P(x_\theta^i),y_\theta)
\end{equation}
where $y_\theta$ is the domain label (0 for source domain and 1 for target domain) and $\mathcal F$ is the Focal Loss\cite{lin2017focal}. By minimizing $L_{dom}$, the network can learn to extract domain-invariant features that are useful for cross-domain detection tasks.

\section{EXPERIMENT}
\subsection{Datasets and Implementation Details}
We conduct extensive experiments on 3 different domain adaptation scenarios, including PASCAL VOC\cite{everingham2010pascal} to Watercolor\cite{inoue2018cross}, PASCAL VOC to Clipart\cite{inoue2018cross} and PASCAL VOC to Comic\cite{inoue2018cross}, where domains with large gaps are evaluated, i.e., from real images to artistic images.

\textbf{Training Details.} The network ResNet-101 pretrained on ImageNet\cite{deng2009imagenet} is used as the backbone according to the settings. We scale all images by resizing the shorter side of the image to 600 while maintaining the image ratios. We first train the networks with only source domain to avoid initial noisy predictions. Then we train the whole model with Adam optimizer and the initial learning rate is set to ${10}^{-5}$, it is divided by 10 at 70, 000, 80, 000 iterations later. The total number of training iterations is 90, 000 and batch size during training is 1. The proposed method is implemented with Pytorch and the networks are trained using one GTX 3080 GPU with 10 GB memory.

\subsection{Comparisons with the State-of-the-Art Methods}
\begin{table*}
	\setlength{\abovecaptionskip}{0pt}
	\centering
	\captionsetup{font={small}}
	\caption{Domain adaptation detection results from PASCAL VOC to Clipart1k}
	\setlength{\tabcolsep}{4.5pt}
	\label{table:label}
	\addtolength{\tabcolsep}{-1.2pt}
	\begin{tabular}{cccccccccccccccccccccc}
		\hline
		Methods& aero & bicy & bird & boat & bott & bus& car& cat& chair& cow& table& dog& hrs& motor& pers& plnt& sheep& sofa& train & tv & mAP \\ \hline
		Source Only\cite{ren2015faster} &4.3&\textbf{75.0}&31.3&21.5 &4.8 &64.5 &23.2 &6.7 &33.1 &7.1 &32.5 &4.9 &56.5 &66.7 &49.6 &51.9 &9.4 &20.5&34.5&21.6 &31.0 \\
		UAN\cite{you2019universal} &33.1&58.9&25.6&26.6&37.7&48.2&31.5&8.6&32.8&23.7&\textbf{31.6}&2.4&36.6&56.6&42.8&44.8&14.7&16.4&35.6&26.5&31.7\\
		CMU\cite{fu2020learning} &32.2&61.1&27.7&14.7&\textbf{41.9}&52.5&34.7&9.2&36.5&38.1&21.0&7.6&37.0&48.6&55.7&44.5&17.7&21.1&35.6&26.5&33.3\\ 
		FBC\cite{yang2020unsupervised}&\textbf{43.9}&64.4&28.9&26.3&39.4&58.9&36.7&\textbf{14.8}&\textbf{46.2}&39.2&11.0&11.0&31.1&\textbf{77.1}&48.1&36.1&17.8&35.2&52.6&50.5&38.5\\
		$I^3$Net\cite{chen2021i3net}&30.0&67.0&32.5&21.8&29.2&62.3&41.3&11.6&	37.1&39.4&27.4&	19.3&25.0&67.4&	55.2&42.9&\textbf{19.5}&\textbf{36.2}&\textbf{50.7}&39.3&37.8\\ 
		UaDaN\cite{guan2021uncertainty}	&35.0&72.7&\textbf{41.0}&24.4&21.3&69.8&\textbf{53.5}&2.3&34.2&	\textbf{61.2}&31&\textbf{29.5}&\textbf{47.9}&63.6&62.2&	\textbf{61.3}&13.9&7.6&48.6&23.9&	\textbf{40.2}\\ \hline
		PDAANet(ours)&38.5&69.8&38.5&\textbf{38.3}&31.8&\textbf{84.4}&43.5&12.3 &43.9&35.1&10.2&15.3&33.1&69.3&\textbf{63.5}&25.8&13.3&19.5&43.4&\textbf{39.4}&38.5(71.9)\\ \hline
	\end{tabular}
\end{table*}
We compare our method with various state-of-the-art methods on three public benchmarks. The results for Clipart, Watercolor, and Comic are reported in Tables \uppercase\expandafter{\romannumeral1}, \uppercase\expandafter{\romannumeral2} and \uppercase\expandafter{\romannumeral3}, respectively. Recall is included in parentheses next to mAP. Our mAP for Clipart is 38.5$\%$, with performance essentially on par with state-of-the-art Methods, and the tasks with the largest accuracy improvements were Bus, Boat, and Person, with improvements of 14.6$\%$, 6.5$\%$, and 1.3$\%$, respectively. For Watercolor, mAP is 57.7$\%$ with a performance improvement of 2.5$\%$, and for Car and Person, it was 3.8$\%$ and 2.0$\%$ respectively. Similarly, mAP on Comic is 34.8$\%$ with a performance improvement of 1.4$\%$, and 1.4$\%$ on Car. Most importantly, our method shows better accuracy than other methods and achieves comparable performance to state-of-the-art methods.

\begin{table}[h]
	\begin{minipage}{\columnwidth}
		\setlength{\belowcaptionskip}{-2pt}
		\centering
		\captionsetup{font={small}}
		\caption{Results from PASCAL VOC to Watercolor2k}
		\setlength{\tabcolsep}{6pt}
		\label{table:label}
		\addtolength{\tabcolsep}{-0.9pt}
		\begin{tabular}{c|ccccccc}
			\hline
			Methods& bike & bird & car & cat & dog & person& mAP\\ \hline
			Source Only\cite{ren2015faster} &68.8 & 46.8 &37.2 &32.7 &21.3 &60.7 &44.6\\
			CMU\cite{fu2020learning}&82.0&53.9&48.6&39.6&33.1&66.0&	53.9\\
			UaDaN\cite{guan2021uncertainty}	&85.6&50.4&40.3&41.5&\textbf{42.3}&68.2&54.7\\
			ATF\cite{he2020domain} &78.8 &\textbf{59.9} &47.9 &41.0 & 34.8 &66.9 &54.9\\
			FBC\cite{yang2020unsupervised} &\textbf{90.1} &49.7 &44.1 &41.1 &34.6 &70.3 &55.0\\
			LGAAD\cite{zhang2021local} &89.5 &52.6 &48.9 &35.1 &37.6 &64.9 &54.8\\ 
			SPANet\cite{fujii2022adversarially} & 81.1 &51.1 &\textbf{53.6} &34.6 &39.8 &71.3 &55.2\\ \hline
			PDAANet(ours)&86.6&52.0&	53.0&\textbf{44.9}&36.1&	\textbf{73.3}&\textbf{57.7}(83.7)\\ \hline
		\end{tabular}
	\end{minipage}
\end{table}
\vspace{-10pt} 
\begin{table}[h]
	\begin{minipage}{\columnwidth}
		\setlength{\abovecaptionskip}{4pt}
		\setlength{\belowcaptionskip}{-2pt}
		\centering
		\captionsetup{font={small}}
		\caption{Results from PASCAL VOC to Comic2k}
		\setlength{\tabcolsep}{6pt}
		\label{table:label}
		\addtolength{\tabcolsep}{-0.9pt}
		\begin{tabular}{c|ccccccc}
			\hline
			Methods & bike & bird & car & cat & dog & person& mAP\\ \hline
			Source Only\cite{ren2015faster}&33.2&14.8&23.8&19.5&19.7 &35.6&24.4\\
			FBC\cite{yang2020unsupervised} &45.7 &17.1 &40.1 &16.5 &22.7 &52.3 &32.4\\
			FGRR\cite{chen2022relation} &42.2 &\textbf{21.1} &30.2 & \textbf{21.9} &30.0 &50.5 &32.7\\
			$I^3$Net\cite{chen2021i3net} &47.5 &19.9 &33.2 &11.4 &19.4 &49.1 &30.1\\
			UaDaN\cite{guan2021uncertainty}	&42.0&12.9&42.6&13.4&14.6&\textbf{61.5}&31.2\\
			AT\cite{fujii2022adversarially} &\textbf{57.1} & 15.4 &41.3 &10.4 &\textbf{23.2} &53.0 &33.4\\ \hline
			PDAANet(ours)&47.4&13.4&\textbf{42.7}&21.0&23.1&61.1&\textbf{34.8}(66.5)\\ \hline
		\end{tabular}
	\end{minipage}
\end{table}

\vspace{-20pt} 
\subsection{Ablation Experiments}
We conducted ablation experiments on the two components of our proposed method, DomMix and PAM, and evaluated the performance of our method on three datasets. Table \uppercase\expandafter{\romannumeral4} shows that DomMix effectively improves detection performance, with the network using mixed features achieving an mAP improvement of about 3$\sim$5. We also evaluated the effectiveness of ECAP and RSA in PAM. The results demonstrate that both of them outperform the single attention module, with mAP improving by 2$\sim$3 to varying degrees.
\vspace{-5pt} 
\begin{table}[h]
	\begin{minipage}{\columnwidth}
		\setlength{\abovecaptionskip}{-0pt}
		\centering
		\captionsetup{font={small}}
		\caption{Ablation Experiments on PDAANet}
		\setlength{\tabcolsep}{12pt}
		\label{tab:mytable}
		\begin{tabular}{c|c|c|c} \hline
			Methods &Watercolor &Comic &Clipart \\ \hline
			Without DomMix	&53.8&29.2&33.1\\ 
			ECAP only&56.2&33.6&35.3\\ 
			RSA only&55.0&33.2&34.7\\ 
			SimAM+ECA &56.6&30.6&32.1\\ \hline
			PDAANet&\textbf{57.7}&\textbf{34.8}&\textbf{36.0}\\ \hline
		\end{tabular}
	\end{minipage}
\end{table}

\vspace{-15pt} 
\subsection{Discussions on the Parameters and Model Effect}
\begin{figure}[h]
	\setlength{\abovecaptionskip}{-0pt}
	\begin{minipage}[t]{0.42\linewidth}
		\centering
		\includegraphics[scale=0.32]{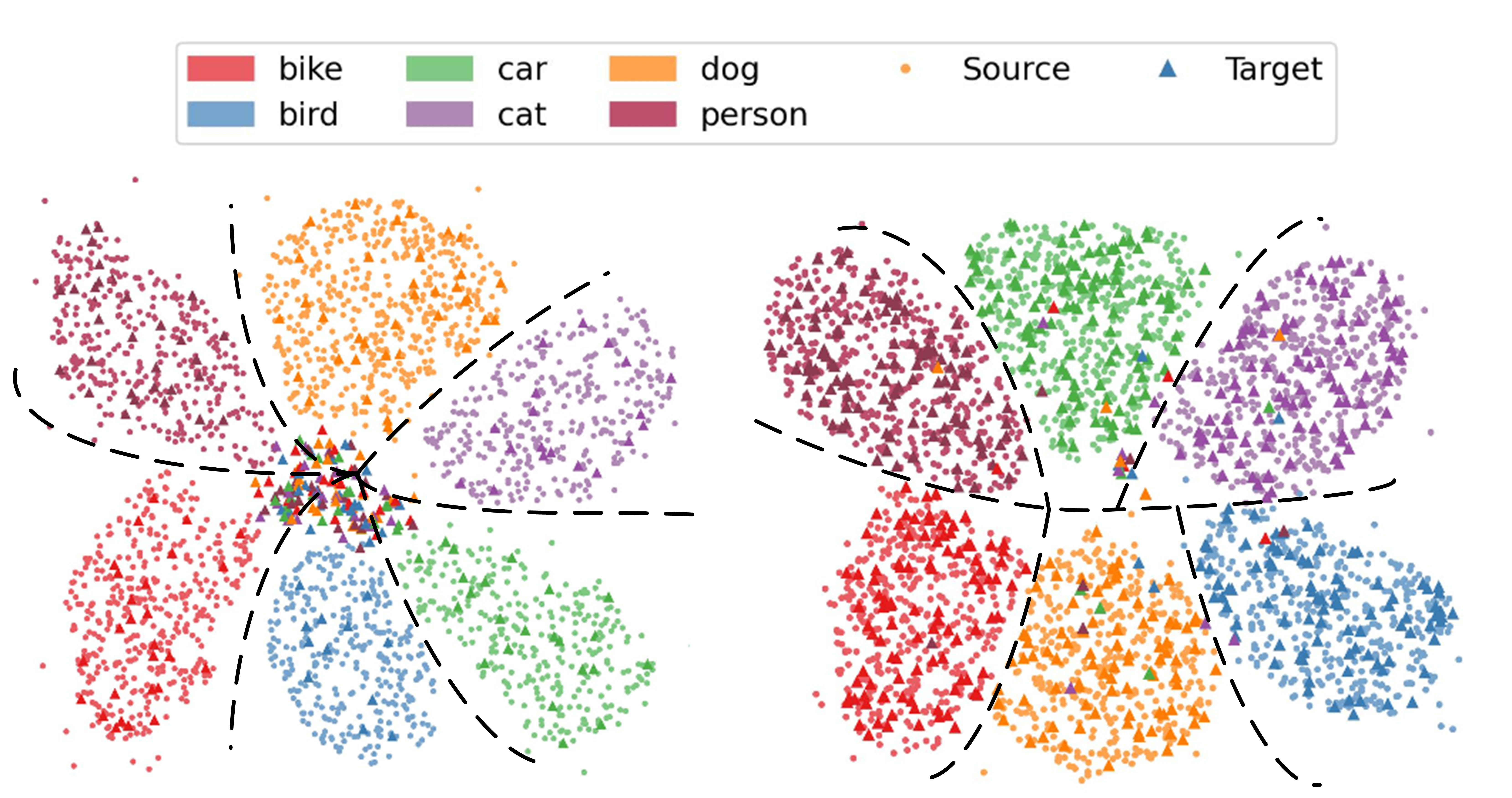}
	\end{minipage}%
	\captionsetup{font={small}}
	\caption{Watercolor2k samples before (left) and after (right) domain adaptation.}
\end{figure}
In Fig. 3, we compared the distributions of target samples before and after domain adaptation. Before adaptation, the target domain samples tend to be clustered together and are difficult to discriminate. After domain adaptation using PAM, the distribution of the target domain samples was well aligned with that of the source domain samples. At the same time, target samples can be well discriminated against, demonstrating the effectiveness of the proposed PAM method.
\begin{figure}[h]
	\begin{minipage}[t]{0.5\linewidth}
		\centering
		\includegraphics[scale=0.3]{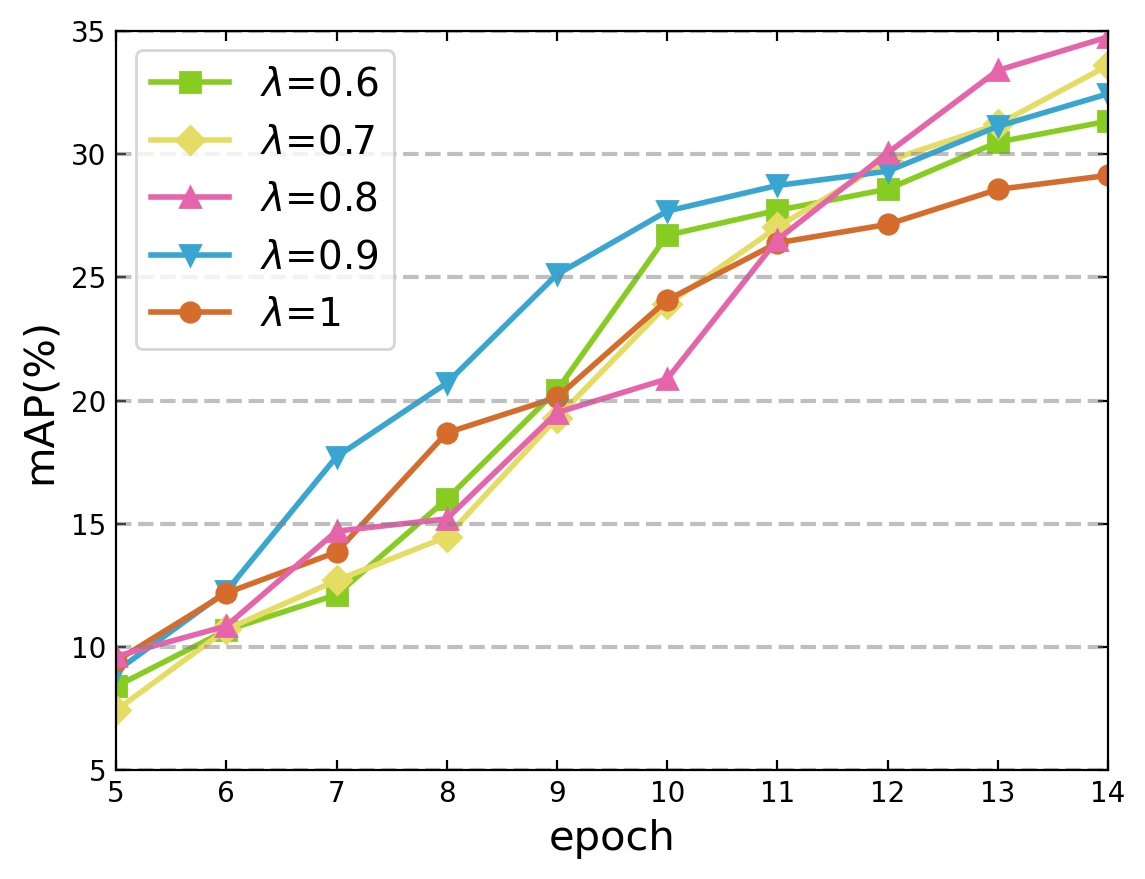}
	\end{minipage}%
	\begin{minipage}[t]{0.5\linewidth}
		\centering
		\includegraphics[scale=0.3]{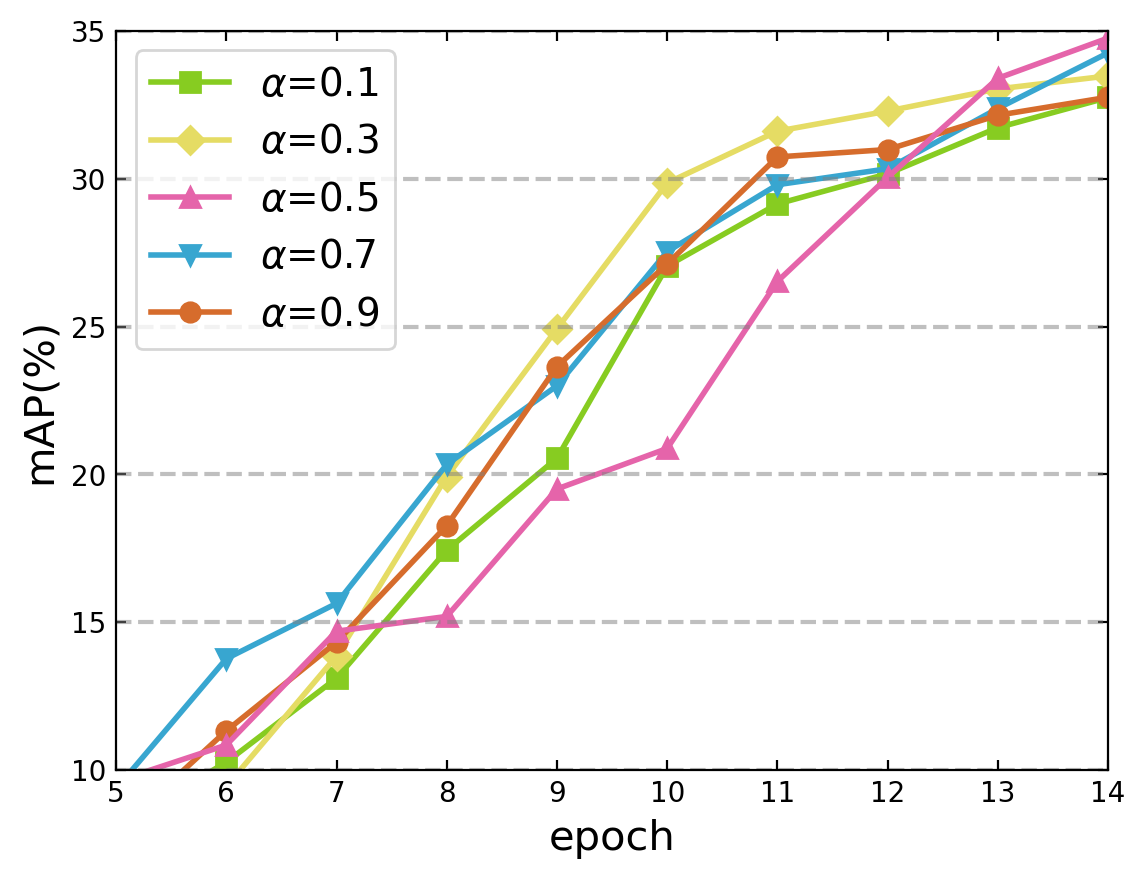}
	\end{minipage}
	\vspace{-10pt}
	\captionsetup{font={small}}
	\caption{Variation of mAP corresponding to different mixing ratios $\lambda$ in DomMix as the number of training rounds increases (left), and variation of mAP corresponding to different weighting factors $\alpha$ in PAM (right).}
\end{figure}

Then we discuss the effect of parameters on PDAANet and use VOC2Comic as an example, as shown in Fig. 4. $\lambda$ represents the mixing ratio of the mixed features in DomMix, with a range from 0.6 to 1, while $\alpha$ represents the weighting coefficient of attention weights in PAM, ranging from 0.1 to 0.9. When $\lambda$ = 1, it means there is no feature mixing between the source domain and target domain. From Fig. 4 (left), the results show that mixing deep features of source domain and target domain features in different proportions can improve training performance to varying degrees. In particular, the best result is obtained when $\lambda$ = 0.8. For the weighting factors $\alpha$ in the PAM, the results show that the maximum of 34.8 can be achieved when $\alpha$ = 0.5, as shown in Fig. 4 (right).

\vspace{-5pt} 
\section{Conclusion}
In this paper, we propose a pairwise DomMix attentive adversarial network for unsupervised DAOD. Driven by a region proposal network, our method constructs a deep mixed intermediate domain by sharing inter-domain differences and optimizes domain alignment through pairwise attention adversarial modules on image-level and instance-level features at different scales. This allows the network to focus on discriminative contextual information between different domains. Extensive experiments on three benchmarks show that our proposed method achieves competitive performance with the state-of-the-art methods.

\newpage
\footnotesize
\bibliography{hyperref}

\end{document}